\documentclass[letterpaper, 10 pt, conference]{ieeeconf} 
\usepackage{times}
\usepackage{color, colortbl, xcolor}
\newcommand{\framework}{SCAFFOLD} 

\newcommand{\pstate}{s} 

\newcommand{\pthresh}{P_{\mathrm{th}}}
\newcommand{\nhumans}{N}

\newcommand{\bq}{\begin{equation}}
\newcommand{\eq}{\end{equation}}
\newcommand{\bmat}{\left[\begin{matrix}}
\newcommand{\emat}{\end{matrix}\right]}

\newcommand{\CA}{robot}
\newcommand{\CAs}{robots}
\newcommand{\NCA}{human}
\newcommand{\NCAs}{humans}

\newcommand{\R}{\mathbb{R}} 

\newcommand{\bqs}{\begin{equation*}}
\newcommand{\eqs}{\end{equation*}}

\def\atan2{\operatorname{atan2}}

\usepackage{multicol}
\usepackage[bookmarks=true]{hyperref}

\usepackage{amsfonts}
\usepackage{algorithmic}
\usepackage[ruled,vlined,titlenotnumbered]{algorithm2e} 
\usepackage{amsmath}
\usepackage{dsfont}
\usepackage{subfigure}
\usepackage{multicol}
\usepackage[pdftex]{graphicx}   
\usepackage{graphicx}

\graphicspath{{./figs/}}    
\DeclareGraphicsExtensions{.pdf,.png}

\usepackage[textfont=md,font=footnotesize]{caption}

\usepackage[left=55pt, right=55pt, top=55pt, bottom=58pt]{geometry} 

\IEEEoverridecommandlockouts
\usepackage{balance}

\usepackage[%
    style=numeric-comp,
    sorting=none,
    backend=biber,
    sortcites=true,
    doi=false,
    firstinits=true,
    hyperref,
    isbn=false,
    eprint=false,
    maxcitenames=3, 
    minbibnames=3, 
    maxbibnames=4, 
    block=none]
    {biblatex}
    
    \renewbibmacro{in:}{}
    \AtEveryBibitem{%
  	\clearlist{language}%
  	\clearfield{pages}%
	}
\bibliography{references}
\usepackage{ifthen}
\newboolean{include-notes}
\setboolean{include-notes}{true}

\newcommand{\adnote}[1]{\ifthenelse{\boolean{include-notes}}{\textcolor{purple}{\textbf{A:#1}}}{}}

\usepackage{lipsum}

\newcommand\blfootnote[1]{%
  \begingroup
  \renewcommand\thefootnote{}\footnote{#1}%
  \addtocounter{footnote}{-1}%
  \endgroup
}

\newcommand{\abnote}[1]%
    {\textcolor{cyan}{\textbf{AB: #1}}}
\newcommand{\jfnote}[1]%
    {\textcolor{orange}{\textbf{JF: #1}}}
\newcommand{\shnote}[1]%
    {\textcolor{blue}{\textbf{SH: #1}}}

\definecolor{planning_color}{RGB}{69, 174, 254}    
\definecolor{prediction_color}{RGB}{255, 116, 190}

\newcommand{\example}[1]%
{
\textbf{Running example:}
\textit{#1}
}

\begin{document}

\title{\quad\\ A Scalable Framework For Real-Time\\ Multi-Robot, Multi-Human Collision Avoidance}

\author{
Andrea Bajcsy*, Sylvia L. Herbert*, David Fridovich-Keil,\\ Jaime F. Fisac, Sampada Deglurkar, Anca D. Dragan, and Claire J. Tomlin
}


%

\maketitle



\begin{abstract}
Robust motion planning is a well-studied problem in the robotics literature, yet current algorithms struggle to operate scalably and safely in the presence of other moving agents, such as humans.
This paper introduces a novel framework for robot navigation that accounts for high-order system dynamics and maintains safety in the presence of external disturbances, other robots, and non-deterministic intentional agents.
Our approach precomputes a tracking error margin for each robot, generates confidence-aware human motion predictions, and coordinates multiple robots with a sequential priority ordering, effectively enabling scalable safe trajectory planning and execution. 
We demonstrate our approach in hardware with two robots and two humans. We also showcase our work's scalability in a larger simulation. 
\end{abstract}

\IEEEpeerreviewmaketitle

\blfootnote{
Department of Electrical Engineering and Computer Sciences,
UC Berkeley,
{\tt \small{\{
abajcsy
\}@berkeley.edu}}.
This research is supported by an NSF CAREER award, the Air Force Office of Scientific Research (AFOSR), NSF's CPS FORCES and VeHICal projects, the UC-Philippine-California Advanced Research Institute, the ONR MURI Embedded Humans, and the SRC CONIX Center.\\
$^*$Indicates equal contribution.
}

\section{Introduction}
\label{sec:intro}



\begin{figure}[t!]
    \centering
    \includegraphics[width=\columnwidth]{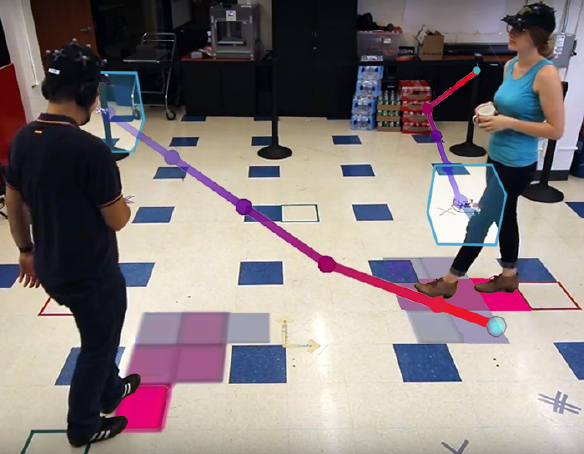}
    \caption{Hardware demonstration of real-time multi-agent planning while maintaining safety with respect to internal dynamics, external disturbances, and intentional \NCAs{}. The planned trajectories from the quadcopters are visualized, and the tracking error bound is shown as a box around each quadcopter. The probabilistic distribution over the future motion of the humans are shown in pink in front of each human.}
    \label{fig:front_fig}
    \vspace{-.6cm}
\end{figure}

As robotic systems are increasingly used for applications such as drone delivery services, semi-automated warehouses, and autonomous cars, safe and efficient robotic navigation around humans is crucial.
Consider the example in Fig.~\ref{fig:front_fig}, inspired by a drone delivery scenario, where two quadcopters must plan a safe trajectory around two humans who are walking through the environment. We would like to guarantee that the robots will reach their goals without ever colliding with each other, any of the humans, or the static surroundings.\footnote{Note that our laboratory setting uses a motion capture system for sensing and state estimation---robustness with respect to sensor uncertainty is an important component that is beyond the scope of this paper.} 

This safe motion planning problem faces three main challenges: (1) controlling the nonlinear robot dynamics subject to external disturbances (e.g. wind), (2) planning around multiple \NCAs{} in real time, and (3) avoiding conflicts with other robots' plans. Extensive prior work from control theory, motion planning, and cognitive science has enabled computational tools 
for
rigorous safety analysis, faster motion planners for nonlinear systems, and predictive models of human agents.
Individually, these problems are difficult---computing robust control policies, coupled robot plans, and joint predictions of multiple human agents are all computationally demanding at best
and intractable at worst~\cite{Mitchell2005, chen2016general}.
Recent work, however, has made progress in provably-safe real-time motion planning~\cite{herbert2017fastrack, majumdar2017funnel, singh2018robust}, real-time probabilistic prediction of a human agent's motion~\cite{fisac2018probabilistically, ziebart2009planning}, and robust sequential trajectory planning for multi-robot systems~\cite{bansal2017safe, chen2016multi}. 
It remains a challenge to synthesize these into a real-time planning system, primarily due to the difficulty of joint planning and prediction for multiple robots and humans. There has been some work combining subsets of this problem \cite{knepper2012pedestrian, trautman2010unfreezing, kruse2013human}, but the full setup of real-time and robust multi-robot navigation around multiple humans remains underexplored.

Our main contributions in this paper are tractable approaches to joint planning and prediction, while still ensuring efficient, probabilistically-safe motion planning. 
\begin{figure*}[!ht]
     \centering
    \includegraphics[width=\textwidth]{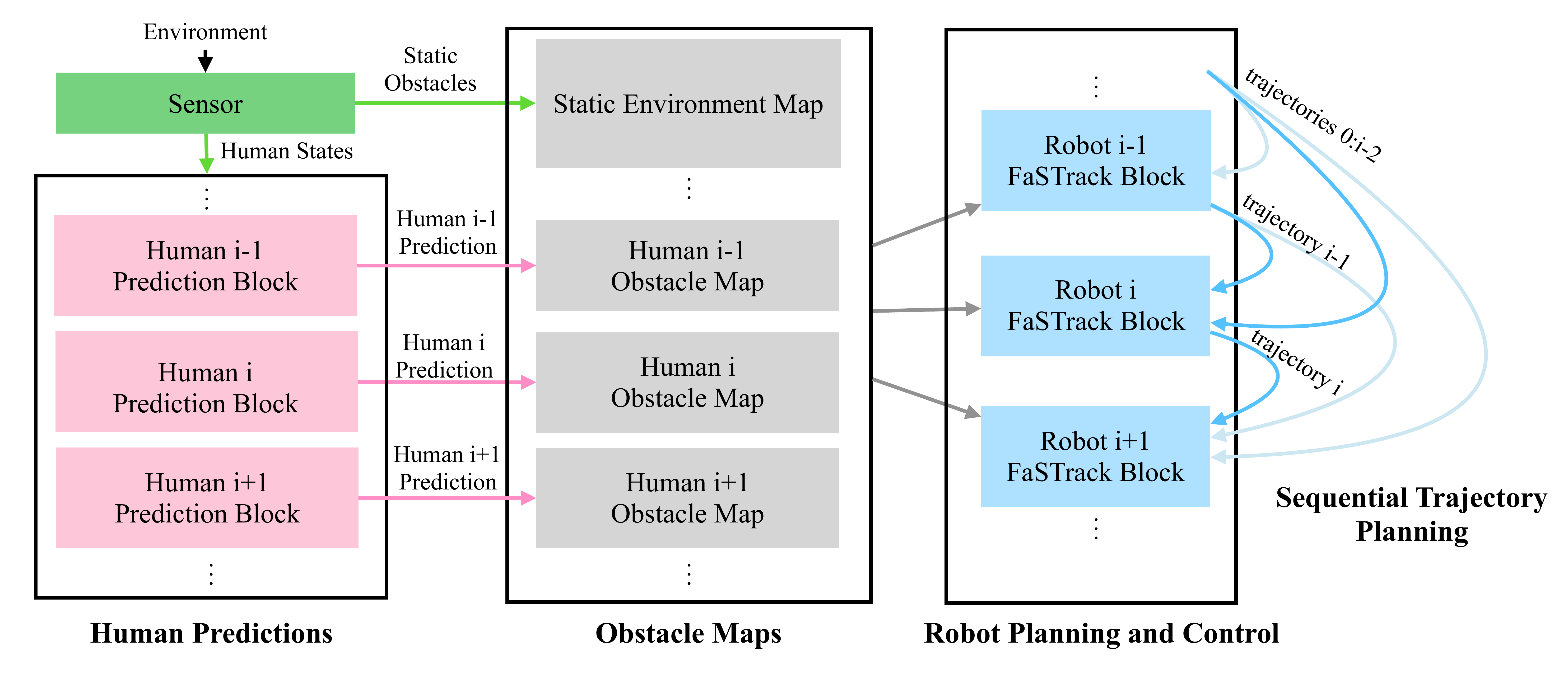}
    \caption{The \framework{} Framework}
    \label{fig:framework}
    \vspace{-.7cm}
\end{figure*}
We use the reachability-based FaSTrack framework \cite{herbert2017fastrack} for real-time robust motion planning.
To ensure real-time feasibility, \CAs{} predict \NCA{} motion using a simple model neglecting future interaction effects.
Because this model will be a simplification of true \NCA{} motion, we use confidence-aware predictions \cite{fisac2018probabilistically} that become more conservative whenever \NCAs{} deviate from the assumed model.
Finally, groups of \CAs{} plan sequentially according to a pre-specified priority ordering \cite{chen2015safe}, which serves to reduce the complexity of the joint planning problem while maintaining safety with respect to each other. 
We demonstrate our framework in hardware, and provide a large-scale simulation to showcase scalability.

\section{The \framework{} Framework}
\label{sec:framework}

Fig. \ref{fig:framework} illustrates our overall planning framework, called SCAFFOLD.
We introduce the components of the framework by incrementally addressing the three main challenges identified above.

We first present the \textbf{robot planning and control} block (Section \ref{sec:fastrack}), which is instantiated for each \CA{}. 
Each \CA{} uses a robust controller (e.g. the reachability-based controller of \cite{herbert2017fastrack}) to track motion plans within a precomputed error margin that accounts for modeled dynamics and external disturbances.
In order to generate safe motion plans, each \CA{} will ensure that output trajectories are collision-checked with a set of obstacle maps, using the tracking error margin. 

These obstacle maps include an \emph{a priori} known set of static obstacles, as well as predictions of the future motion of any \NCAs{}, which are generated by the \textbf{human predictions} block (Section \ref{sec:predictions}). 
By generating these predictions, each \CA{} is able to remain probabilistically safe with respect to the \NCAs{}. 
To ensure tractability for multiple humans, we generate predictions using simplified interaction models, and subsequently adapt them
following a real-time Bayesian approach such as~\cite{fisac2018probabilistically}.
We leverage the property that individual predictions automatically become more uncertain whenever their accuracy degrades, and use this to enable our tractable predictions to be robust to unmodeled interaction effects.

Finally, to guarantee safety with respect to other \CAs, we carry out \textbf{sequential trajectory planning} (Section \ref{sec:STP}) by adapting the cooperative multi-agent planning scheme~\cite{bansal2017safe}
to function in real time with the robust trajectories from the planning and control block.
The \CAs{} generate plans according to a pre-specified priority ordering. Each \CA{} plans to avoid the most recently generated trajectories from \CAs{} of higher priority, i.e. \CA{} $i$ must generate a plan that is safe with respect to the planned trajectories from \CAs{} $j, j < i$. 
This removes the computational complexity of planning in the joint state space of all \CAs{} at once.

\section{Robot Planning and Control}
\label{sec:fastrack}

In this section we begin with the canonical problem of planning through a static environment.
Efficient algorithms such as A$^*$ or rapidly-exploring random trees (RRT) \cite{hart1968astar, karaman2011RRTPRM} excel at this task. 
These algorithms readily extend to environments with deterministically-moving obstacles by collision-checking in both time and space.

We now introduce \CA{} dynamics and allow the environment to have external disturbances such as wind.
Kinematic planners such as A$^*$ and RRT do not consider these factors when creating plans. 
In practice, however, these planners are often used to generate an initial trajectory, which may then be smoothed and tracked using a feedback controller such as a linear quadratic regulator (LQR). 
During execution, the mismatch between the planning model and the physical system can result in tracking error, which may cause the \CA{} to deviate far enough from its plan to collide with an obstacle. 
To reduce the chance of collision, one can augment the \CA{} by a heuristic error buffer; this induces a ``safety bubble'' around the \CA{} used when collision checking. However, heuristically generating this buffer will not guarantee safety.

Several recent approaches address efficient planning while considering model dynamics and maintaining robustness with respect to external disturbances. Majumdar and Tedrake~\cite{majumdar2017funnel} use motion primitives with safety funnels, while Rakovi{\'c}~\cite{rakovic2009set} utilizes robust model-predictive control, and Singh et. al.~\cite{Singh2017} leverage contraction theory.  

\begin{figure}
    \centering
    \includegraphics[width=.7\columnwidth]{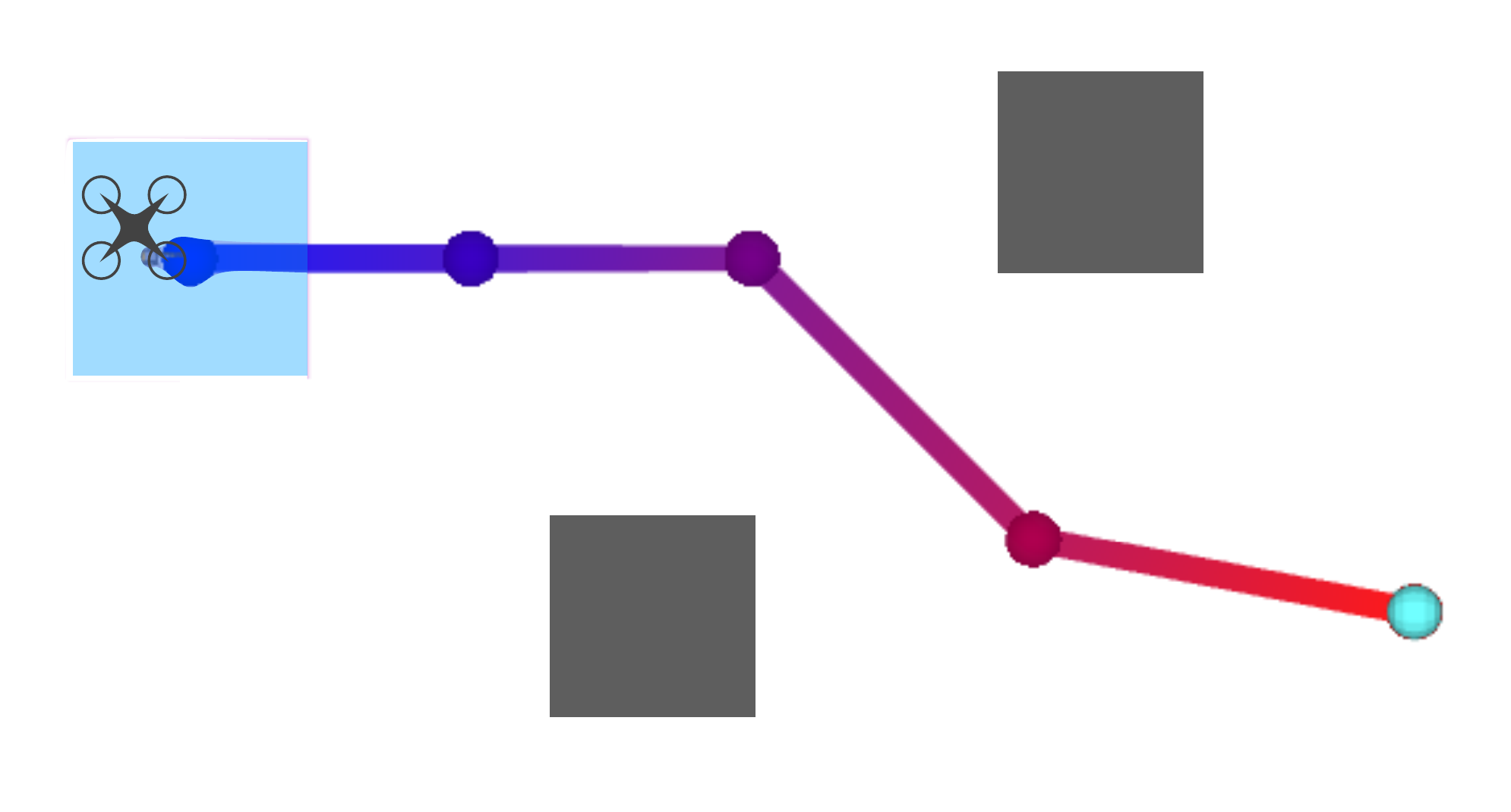}
    \caption{Top-down view of FaSTrack applied to a 6D quadcopter navigating a static environment. Note the simple planned trajectory (changing color over time) and the tracking error bound (TEB) around the quadcopter. This TEB is a 6D set that has been projected down to the position dimensions. Because we assuem the quadcopter moves independently in $(x,y,z)$, this projection looks like a box, making collision-checking very straightforward.}
    \label{fig:fastrack_birds_eye}
    \vspace{-.5cm}
\end{figure}

In this paper, we use FaSTrack \cite{herbert2017fastrack, fridovich2018planning}, a modular framework that  computes a tracking error bound (TEB) via \emph{offline} reachability analysis. 
This TEB can be thought of as a rigorous counterpart 
of the error-buffer concept introduced above. 
More concretely, the TEB is the set of states capturing the maximum relative distance (i.e. maximum tracking error) that may occur between the physical \CA{} and the current state of the planned trajectory.
We compute the TEB by formulating the tracking task as a pursuit-evasion game between the planning algorithm and the physical robot. We then solve this differential game using Hamilton-Jacobi reachability analysis. To ensure robustness, we assume (a) worst-case behavior of the planning algorithm (i.e. being as difficult as possible to track), and (b) that the \CA{} is experiencing worst-case, bounded external disturbances.
The computation of the TEB also provides a corresponding error-feedback controller for the robot to always remain inside the TEB. 
Thus, FaSTrack wraps efficient motion planners, and adds robustness to modeled system dynamics and external disturbances through the precomputed TEB and error-feedback controller. 
Fig. \ref{fig:fastrack_birds_eye} shows a top-down view of a quadcopter using a kinematic planner (A$^*$) to navigate around static obstacles. By employing the error-feedback controller, the quadcopter is guaranteed to remain within the TEB (shown in blue) as it traverses the A$^*$ path.

\subsection{FaSTrack Block}
\textbf{\textit{Requirements:}} To use FaSTrack, one needs a high-fidelity dynamical model of the system used for reference tracking, and a (potentially simpler) dynamic or kinematic model used by the planning algorithm.
Using the relative dynamics between the tracking model and the planning model, the TEB and safety controller may be computed using Hamilton-Jacobi reachability analysis \cite{herbert2017fastrack}, sum-of-squares optimization \cite{singh2018robust}, or approximate dynamic programming \cite{royo2018classification}. 

\begin{figure}
    \centering
    \includegraphics[width=.9\columnwidth]{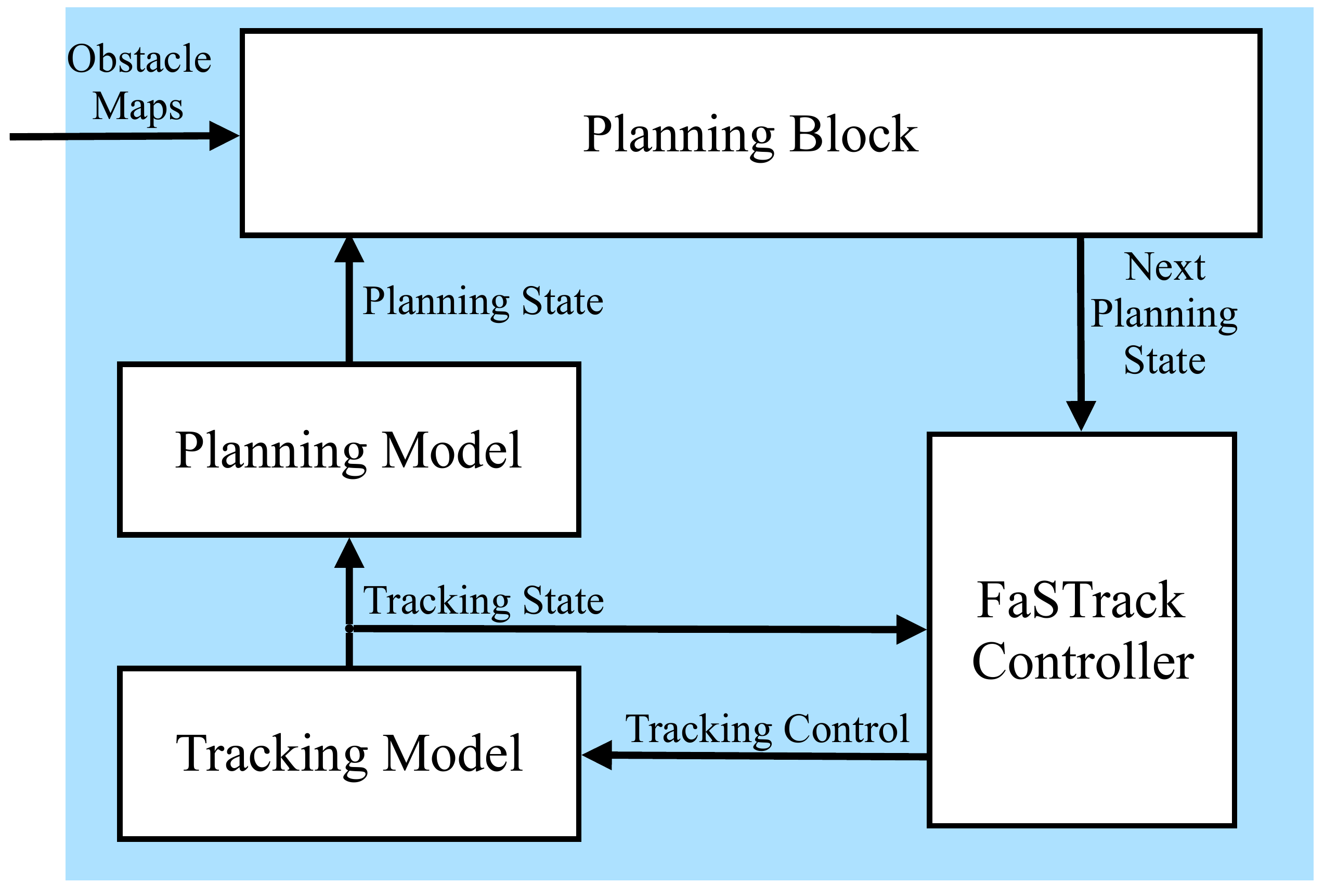}
    \caption{FaSTrack Block}
    \label{fig:fastrack}
    \vspace{-.3cm}
\end{figure}

\textbf{\textit{Implementation:}}
Fig.~\ref{fig:fastrack} describes the online algorithm for FaSTrack after the offline precomputation of the TEB and safety controller. We initialize the 
\textbf{planning block} 
to start within the TEB centered on the \CA{}'s current state.  The planner then uses any desired planning algorithm (e.g. A$^*$, or model predictive control) to find a trajectory from this initial state to a desired goal state. 
When collision-checking, the planning algorithm must ensure that the tube defined by the Minkowski sum of the TEB and the planned trajectory does not overlap any obstacles in the \textbf{obstacle map}. 

The planning block provides the current planned reference state to the \textbf{FaSTrack controller}, which determines the relative state between the tracking model (\CA{}) and planned reference (motion plan). The controller then applies the corresponding optimal, safe tracking control via an efficient look-up table. 

\subsection{FaSTrack in the \framework{} Framework}
In the \textbf{robot planning and control} section of Fig.~\ref{fig:framework}, each \CA{} uses FaSTrack for robust planning and control. 
FaSTrack guarantees that each \CA{} remains within its TEB-augmented trajectory.

\section{Human Predictions}
\label{sec:predictions}

Section~\ref{sec:fastrack} introduced methods for the fast and safe navigation of a single \CA{} in an environment with deterministic, moving obstacles. 
However, moving obstacles---especially human beings---are not always best modeled as deterministic. 
For such ``obstacles,'' \CAs{} can employ probabilistic predictive models to produce a distribution of states the \NCA{} may occupy in the future. The quality of these predictions and the methods used to plan around them determine the overall safety of the system.
Generating accurate real-time predictions for multiple humans (and, more generally, uncertain agents) is an open problem.
Part of the challenge arises from the combinatorial explosion of interaction effects as the number of agents increases.
Any simplifying assumptions, such as neglecting interaction effects,
will inevitably cause predictions to become inaccurate.
Such inaccuracies could threaten the safety
of plans that rely on these predictions.

Our goal is to compute real-time motion plans that are based on up-to-date predictions of all \NCAs{} in the environment, and at the same time maintain safety when these predictions become inaccurate.
The confidence-aware prediction approach of \cite{fisac2018probabilistically} provides a convenient mechanism for adapting prediction uncertainty online to reflect the degree to which \NCAs{}' actions match an internal model.
This automatic uncertainty adjustment allows us to simplify or even neglect interaction effects between \NCAs{}, because uncertain predictions will automatically result in more conservative plans when the observed behavior departs from internal modeling assumptions.

\subsection{Human Prediction Block}

\textbf{\textit{Requirements:}}
In order to make any sort of collision-avoidance guarantees, we require a prediction algorithm that produces distributions over future states, and rapidly adjusts those predictions such that the actual trajectories followed by \NCAs{} lie within the prediction envelope. There are many approaches to probabilistic trajectory prediction in the literature, e.g. \cite{ding2011human,hawkins2013probabilistic, ziebart2009planning, lasota2015analyzing}. 
These methods could be used to produce a probabilistic prediction of the $i$-th human's state $x_i\in\R^{n_i}$ at future times $\tau$, conditioned on observations%
\footnote{For simplicity, we will later assume complete state observability: ${z^t=x^t}$. 
}
$z$: $P(x_i^\tau | z^{0:t})$. These observations are random variables and depend upon the full state of all \CAs{} and \NCAs{} $x$ until the current time $t$. However, by default these distributions will not necessarily capture the true trajectories of each \NCA{}, especially when the models do not explicitly account for interaction effects. Fisac et. al.~\cite{fisac2018probabilistically} provide an efficient mechanism for updating the uncertainty (e.g., the variance) of distributions $P(x_i^\tau | z^{0:t})$ to satisfy this safety requirement.

\begin{figure}
    \centering
    \includegraphics[width=\columnwidth]{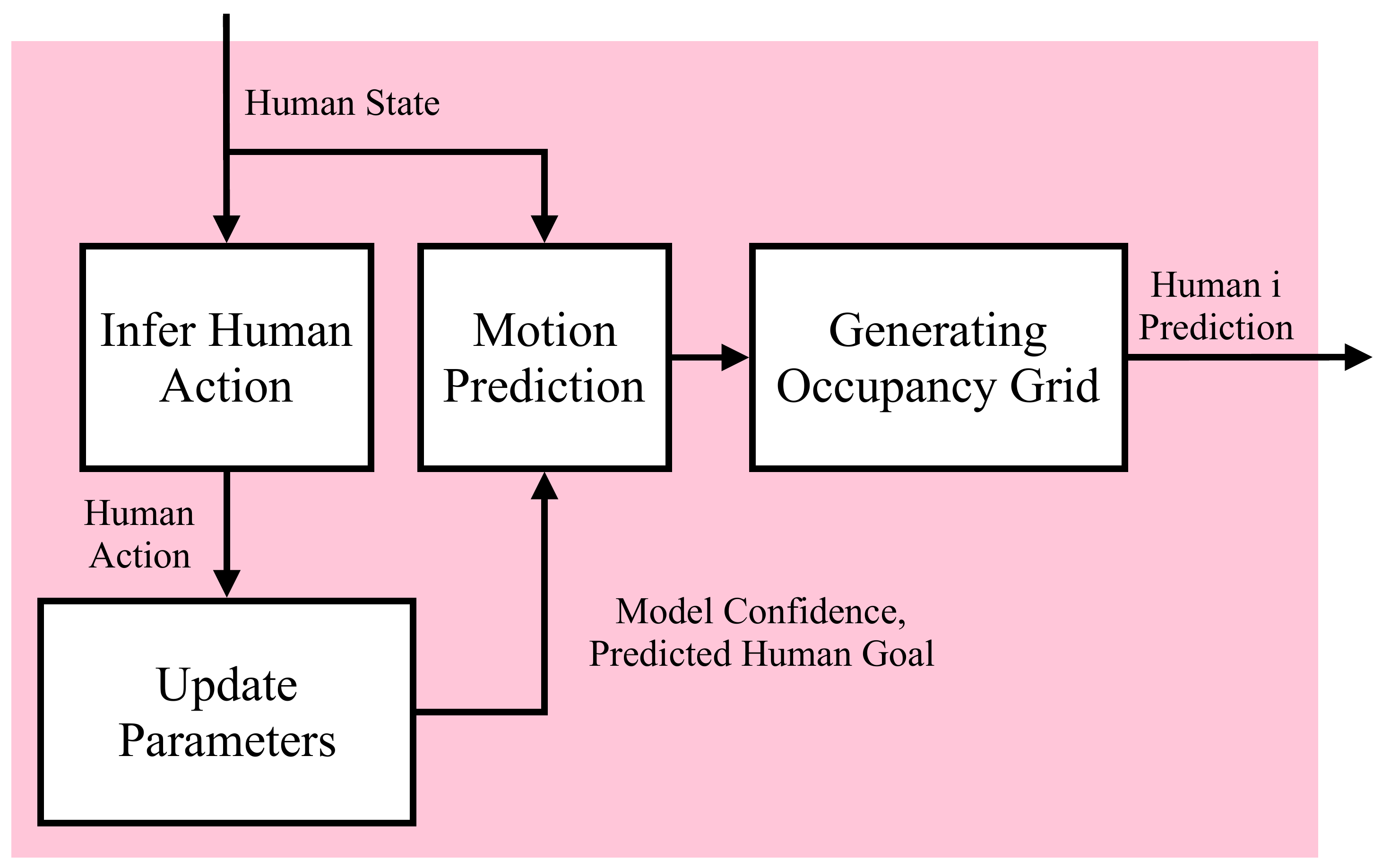}
    \caption{Human Prediction Block}
    \label{fig:prediction}
    \vspace{-.4cm}
\end{figure}

\textbf{\textit{Implementation:}}
Fig. \ref{fig:prediction} illustrates the \textbf{human prediction block}. We use a maximum-entropy model 
as in \cite{fisac2018probabilistically, finn2016guided, ziebart2008maximum}, in which the dynamics of the $i$-th \NCA{} are affected by actions $u_i^t$ drawn from a Boltzmann probability distribution. This time-dependent distribution over actions implies a distribution over future states.
Given a sensed state $x_i^t$ of \NCA{} $i$ at time $t$, we invert the dynamics model to infer the \NCA{}'s action, $u_i^t$.
Given this action, we perform a Bayesian update on the distribution of two parameters: $\theta_i$, which describes the objective of the \NCA{} (e.g. the set of candidate goal locations), and $\beta_i$, which
governs the variance of the predicted action distributions.
$\beta_i$ can be interpreted as a natural
indicator of \textit{model confidence}, quantifying the model's ability to capture \NCAs{}' current behavior~\cite{fisac2018probabilistically}.
Were we to model actions with a different distribution, e.g. a Gaussian process, the corresponding parameters could be learned from prior data \cite{ziebart2008maximum,ziebart2009planning,finn2016guided}, or inferred online \cite{Sadigh2016information,fisac2018probabilistically} using standard inverse optimal control (inverse reinforcement learning) techniques.

\begin{figure}
    \centering
    \includegraphics[width=.6\columnwidth]{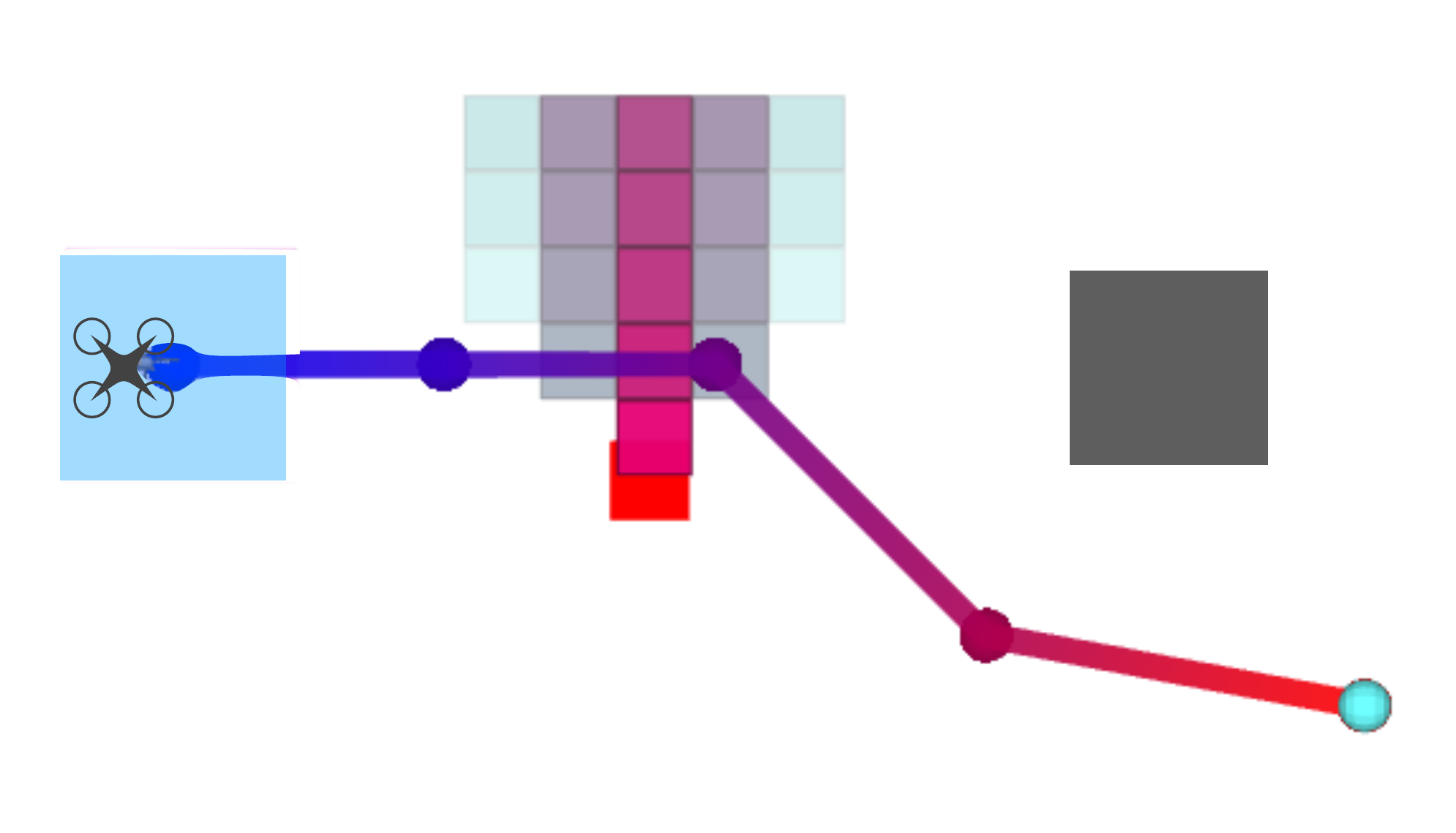}
    \caption{Our environment now has a \NCA{} (red square). The robot models the \NCA{} as likely to move north. Visualized on top of the \NCA{} is the distribution of future states (pink is high, blue is low probability). Since the \NCA{} is walking north and matching the model, the \CA{}'s predictions are confident that the \NCA{} will continue northward and remain collision-free. 
    }
    \label{fig:humans_birds_eye}
     \vspace{-.6cm}
\end{figure}

Once distributional parameters are updated, we produce a prediction over the future actions of \NCA{} $i$ through the following Boltzmann distribution:
\begin{equation}\label{eq:Boltzmann}
    P(u^t_i \mid x^t; \beta_i, \theta_i) \propto e^{\beta_i Q_i(x^t,u^t_i; \theta_i)}\enspace.
\end{equation}
This model treats each \NCA{} as more likely to choose actions with high expected utility as measured by the (state-action) Q-value
associated to a certain reward function, $r_i(x,u_i; \theta_i)$. In general, this value function may depend upon the joint state $x$ and the \NCA{}'s own action $u_i$, as well as the parameters $\theta_i, \beta_i$.
Finally, combining~\eqref{eq:Boltzmann} with a dynamics model, these predicted actions may be used to generate a distribution over future states. In practice, we represent this distribution as a discrete occupancy grid. One such grid is visualized in Fig.~\ref{fig:humans_birds_eye}.

By reasoning about each \NCA{}'s model confidence as a hidden state \cite{fisac2018probabilistically}, our framework dynamically adapts predictions
to the evolving accuracy of the models encoded in the set of state-action functions, $\{Q_i\}$.
Uncertain predictions will force the planner to be more cautious whenever the actions of the \NCAs{} occur with low probability as measured by \eqref{eq:Boltzmann}.


\subsection{Human Prediction in the \framework{} Framework}
The predicted future motion of the \NCAs{} is generated as a probability mass function,
represented as time-indexed set of occupancy grids.
These distributions are interpreted as an obstacle map by the FaSTrack block. 
During planning, a state is considered to be unsafe if the total probability mass contained within the TEB centered at that state exceeds a preset threshold, $\pthresh$. As in \cite{fisac2018probabilistically}, we consider a trajectory to be unsafe if the maximum \emph{marginal} collision probability at any individual state along it exceeds $\pthresh$.

When there are multiple \NCAs{}, their state at any future time $\tau$ will generally be characterized by a joint probability distribution $P(x_{1}^\tau,...,x_{\nhumans}^\tau)$.

Let $\pstate^\tau$ be the planned state of a \CA{} at time $\tau$. We write $\mathrm{coll}(\pstate^\tau, x_i^\tau)$ to denote the overlap of the TEB centered at $\pstate^\tau$ with the $i$th \NCA{} at state $x_i^\tau$. Thus, we may formalize the probability of collision with \emph{at least one} \NCA{} as:
\begin{align}\label{eq:joint_coll_probability}
    P\big(\text{coll}&(\pstate^\tau,\{x_i^\tau\}_{i=1}^\nhumans)\big) =  \\
    &1 - \prod_{i=1}^\nhumans P\big(\neg\text{coll}(\pstate^\tau,x_i^\tau)\mid
                \neg\text{coll}(\pstate^\tau,\{x_j^\tau\}_{j=1}^{i-1})\big)
    \enspace,\notag
\end{align}
%

Intuitively, \eqref{eq:joint_coll_probability} states that the probability that the robot is in collision at $s^\tau$ is one minus the probability that the robot is not in collision. We compute the second term by taking the product over the probability that the robot is not in collision with each human, given that the robot is not in collision with all previously accounted for humans. Unfortunately, it is generally intractable to compute the terms in the product in \eqref{eq:joint_coll_probability}.
Fortunately, tractable approximations can be computed by storing only the marginal predicted distribution of each human at every future time step $\tau$, and assuming independence between humans.
This way, each \CA{} need only operate with $\nhumans$ occupancy grids. The resulting computation is: \vspace{-.2cm}
\begin{equation}\label{eq:marginal_coll_probability}
    P\big(\text{coll}(\pstate^\tau,\{x_i^\tau\}_{i=1}^\nhumans)\big) \approx
    1 - \prod_{i=1}^\nhumans \Big(1-P\big(\text{coll}(\pstate^\tau,x_i^\tau)\big)\Big)
    \enspace.
\end{equation}
%
Here we take the product over the probability that the robot is not in collision with each human (one minus probability of collision), and then again take the complement to compute the probability of collision with any human. Note that when the predictive model neglects future interactions between multiple \NCAs{}, \eqref{eq:joint_coll_probability} reduces to \eqref{eq:marginal_coll_probability}.
If model confidence analysis \cite{fisac2018probabilistically} is used in conjunction with such models, we hypothesize that each marginal distribution will naturally become more uncertain when interaction effects are significant.

Once a collision probability is exactly or approximately computed, the planner can reject plans for which, at any time $\tau>t$, 
the probability of collision from \eqref{eq:marginal_coll_probability} exceeds $\pthresh$. Thus, we are able to generate computationally tractable predictions that result in \mbox{$\pthresh$-safe} planned trajectories for the physical robot.

 \begin{figure*}[htbp]
     \centering
    \includegraphics[width=.9\textwidth]{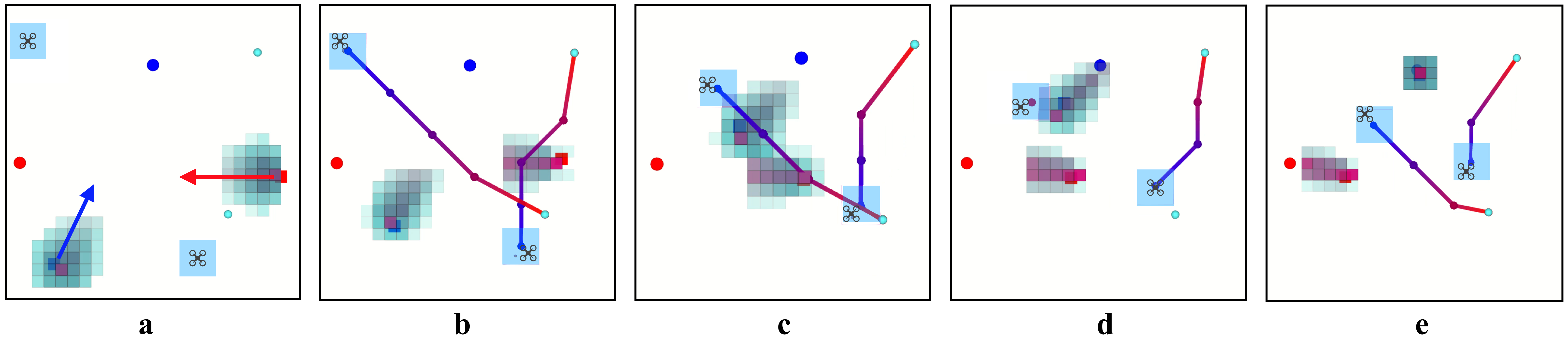}
    \caption{Birds-eye Robot Operating System (ROS) visualization of hardware demonstration from Fig. \ref{fig:front_fig}. (a) Two humans (red and blue) start moving towards their respective goals (also red and blue). Robot in lower right-hand corner has first priority, and robot in upper left-hand corner has second. The time-varying predictions of each human's future motion are visualized. (b) Robots plan trajectories to their goals based on the predictions, priority order, and are guaranteed to stay within the tracking error bound (shown in blue). (c) When the humans begin to interact in an unmodeled way by moving around each other, the future predictions become more uncertain. (d) The robots adjust their plans to be more conservative--note the upper-left robot waiting as the blue human moves past. (c) When the humans pass each other and the uncertainty decreases, the robots complete their trajectories.}.
    \label{fig:robot_planning_algorithm}
    \vspace{-1.0cm}
\end{figure*}

\section{Sequential Trajectory Planning}
\label{sec:STP}

Thus far, we have shown how our framework allows a single \CA{} to navigate in real-time through an environment with multiple \NCAs{} while maintaining safety (at a probability of approximately \mbox{$\pthresh$-safe}) and accounting for internal dynamics, external disturbances, and \NCAs{}. 
However, in many applications (such as autonomous driving), the environment may also be occupied by other \CAs{}.


Finding the optimal set of trajectories for all \CAs{} in the environment would require solving the planning problem over the joint state space of all \CAs{}.  This very quickly becomes computationally intractable with increasing numbers of \CAs{}. 
Approaches for multi-robot trajectory planning often assume that the other vehicles operate with specific control strategies such as those involving induced velocity obstacles \cite{wu2012guaranteed, fiorini1998motion, chasparis2005linear, van2008reciprocal} and involving virtual structures or potential fields to maintain collision \cite{olfati2002distributed, chuang2007multi, zhou2018agile}.
These assumptions greatly reduce the dimensionality of the problem, but may not hold in general.

Rather than assuming specific control strategies of other \CAs{}, each \CA{} could generate predictions over the future motion of all other \CAs{}.
Successful results of this type typically assume that the vehicles operate with very simple dynamics, such as 
single integrator dynamics \cite{Zhou2017}, differentially flat systems \cite{lian2002real}, linear systems \cite{ahmadzadeh2009multi}.

However, when \CAs{} can communicate with each other, methods for centralized and/or cooperative multi-agent planning allow for techniques for scalability \cite{lewis2013cooperative, torreno2017cooperative, mylvaganam2017differential}.
One such method is sequential trajectory planning (STP) \cite{chen2015safe}, which coordinates robust multi-agent planning using a sequential priority ordering.
Priority ordering is commonly used in many multi-agent scenarios, particularly for aerospace applications.
In this work, we merge STP with FaSTrack to produce real-time planning for multi-agent systems.

\subsection{Sequential Trajectory Planning}
\textbf{\textit{Requirements:}} To apply STP, \CAs{} must be able to communicate trajectories and TEBs over a network.

\textbf{\textit{Implementation:}}
STP addresses the computational complexity of coupled motion planning by assigning a priority order to the \CAs{} and allowing higher-priority \CAs{} to ignore the planned trajectories of lower-priority \CAs{}. 
The first-priority \CA{} uses the \textbf{FaSTrack block} to plan a (time-dependent) trajectory through the environment while avoiding the \textbf{obstacle maps}. 
This trajectory is shared across the network with all lower-priority \CAs{}.
The $i$-th \CA{} augments the trajectories from \CAs{} $0:i-1$ by their respective TEBs, and treats them as time-varying obstacles. 
The $i$-th \CA{} determines a safe trajectory that avoids these time-varying tubes as well as the predicted state distributions of \NCAs{}, and publishes this trajectory for \CAs{} $i+1:n$.
This process continues until all \CAs{} have computed their trajectory.  
Each \CA{} replans as quickly as it is able; in our experiments, this was between $50$--$300$ ms.

\subsection{Sequential Trajectory Planning in the \framework{} Framework}
By combining this method with FaSTrack for fast individual planning, the sequential nature of STP does not significantly affect overall planning time. In our implementation all computations are done on a centralized computer using the Robot Operating System (ROS), however this method can easily be performed in a decentralized manner. Note that STP does depend upon reliable, low-latency communication between the \CAs{}. If there are communication delays, techniques such as \cite{desai2017drona} may be used to augment each \CA{}'s TEB by a term relating to time delay.

\section{Implementation and Experimental Results}
\label{sec:demo}

We demonstrate \framework{}'s feasibility in hardware with two \CA{}s and two \NCA{}s, and its scalability in simulation with five \CA{}s and ten \NCA{}s.

\subsection{Hardware Demonstration}
We implemented the \framework{} framework in C++ and Python, using Robot Operating System (ROS) \cite{quigley2009ros}. All computations for our hardware demonstration were done on a laptop computer (specs: 31.3 GB of memory, 216.4 GB disk, Intel Core i7 @ 2.70GHz x 8). As shown in Fig.~\ref{fig:front_fig}, we used Crazyflie 2.0 quadcopters as our \CAs, and two \NCA{} volunteers. The position and orientation of \CAs{} and \NCAs{} were measured at roughly 235 Hz by an OptiTrack infrared motion capture system.  
The \NCA{}s were instructed to move towards different places in the lab, while the quadcopters planned collision-free trajectories in three dimensions $(x,y,z)$ using a time-varying implementation of A$^*$. 
The quadcopters tracked these trajectories using the precomputed FaSTrack controller designed for a 6D near-hover quadcopter model tracking a 3D point \cite{fridovich2018planning}. Human motion was predicted $2$ s into the future. Fig.~\ref{fig:robot_planning_algorithm} shows several snapshots of this scene over time. Note that the humans must move around each other to reach their goals---this is an unmodeled interaction affect.  The predictions become less certain during this interaction, and the quadcopters plan more conservatively, giving the humans a wider berth.  The full video of the hardware demonstration can be viewed in our video submission.



\subsection{\framework{} Framework Simulation Analysis}
Due to the relatively small size of our motion capture arena, we demonstrate scalability of the \framework{} framework through a large-scale simulation. 
In this simulation, pedestrians are crossing through a $25 \times 20 \textrm{m}^2$ region of the UC Berkeley campus.
We simulate the pedestrians' motion using potential fields \cite{goodrich2002potential}, which ``pull'' each pedestrian toward his or her own goal and ``push'' them away from other pedestrians and \CAs{}.
These interaction effects between \NCAs{} and \CAs{} are not incorporated into the state-action functions $\{Q_i\}$, and lead to increased model uncertainty (i.e., higher estimates of $\beta_i$) during such interactions.
The fleet of robots must reach their respective goals while maintaining safety with respect to their internal dynamics, humans, and each other. We ran our simulation on a desktop workstation with an Intel i7 Processor and 12 CPUs operating at 3.3 GHz.%
\footnote{The computation appears to be dominated by the prediction step, which we have not yet invested effort in optimizing.}
Our simulation took $98$ seconds for all robots to reach their respective goals. Predictions over human motion took $0.15 \pm 0.06$ seconds to compute for each human.  This computation can be done in parallel. Each robot took $0.23 \pm 0.16$ seconds to determine a plan. There was no significant difference in planning time between robots of varying priority.
Robots used time-varying A$^*$ on a $2$-dimensional grid with $1.5$ m resolution, and collision checks were performed at $0.1$ m along each trajectory segment. The resolution for human predictions was $0.25$ m and human motion was predicted $2$ s into the future.

\begin{figure}
    \centering
    \includegraphics[width=.8\columnwidth]{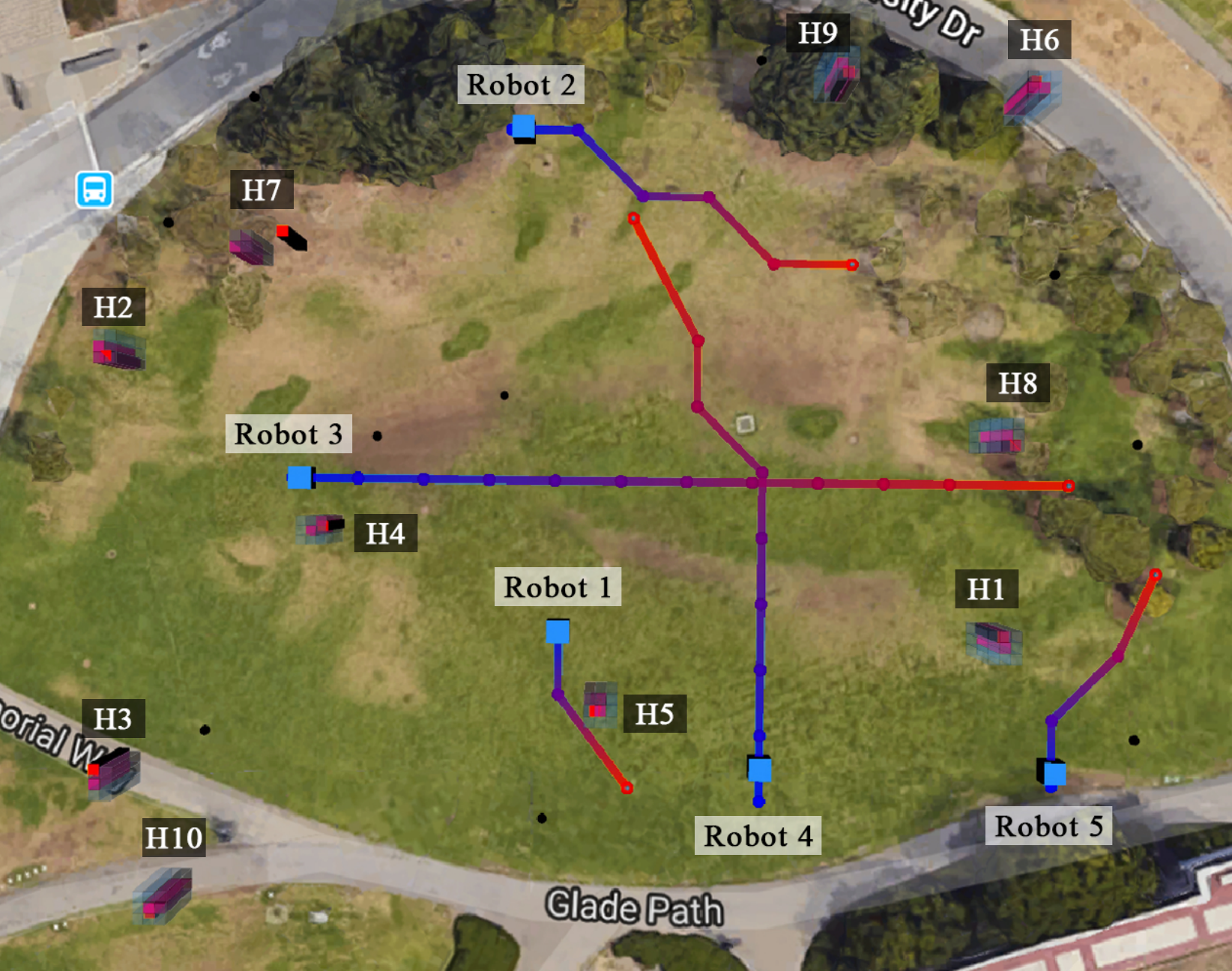}
    \caption{Simulation of 5 dynamic robots navigating in a scene with 10 humans. The simulated humans according to a potential field, which results in unmodeled interaction effects. However, \framework{} enables each robot to still reach its goal safely.}
    \label{fig:simulation}
    \vspace{-.7cm}
\end{figure}


\section{Discussion \& Conclusion}
\label{sec:conclusion}
In this paper, we compose several techniques for robust and efficient planning together in a framework designed for fast multi-\CA{} planning in environments with uncertain moving obstacles, such as \NCAs{}. 
Each \CA{} generates real-time motion plans while maintaining safety with respect to external disturbances and modeled dynamics via the FaSTrack framework.
To maintain safety with respect to \NCAs{}, \CAs{} sense \NCAs{}' states and form probabilistic, adaptive predictions over their future trajectories. For efficiency, we model these \NCAs{}' motions as independent, and to maintain robustness, we adapt prediction model confidence online.
Finally, to remain safe with respect to other \CAs{}, we introduce multi-\CA{} cooperation through STP, which relieves the computational complexity of planning in the joint state space of all \CAs{} by instead allowing \CAs{} to plan sequentially according to a fixed priority ordering.

We demonstrate our framework in hardware with two quadcopters navigating around two \NCAs{}. We also present a larger simulation of five quadcopters and ten \NCAs{}.

To further demonstrate our framework's robustness, we are interested in exploring (a) non-grid based methods of planning and prediction, (b) the incorporation of sensor uncertainty, (c) optimization for timing and communication delays, and (d) recursive feasibility in planning. We are also interested in testing more sophisticated predictive models for \NCAs{}, and other low-level motion planners.

\section*{Acknowledgments}
The authors would like to thank Joe Menke for his assistance in hardware and motion capture systems, and Daniel Hua and Claire Dong for their help early on. 

\balance
\printbibliography
\end{document}